\documentclass[sigconf, screen]{acmart}

\usepackage{amsmath}
\usepackage{amsfonts}
\usepackage{graphicx}
\usepackage{xcolor}
\usepackage{soul}
\usepackage{subcaption}
\usepackage{url}
\usepackage{hyperref}
\AtBeginDocument{%
  }

\copyrightyear{2025}
\acmYear{2025}
\setcopyright{cc}
\setcctype{by}
\acmConference[SIGSPATIAL '25]{The 33rd ACM International Conference on
Advances in Geographic Information Systems}{November 3--6,
2025}{Minneapolis, MN, USA}
\acmBooktitle{The 33rd ACM International Conference on Advances in
Geographic Information Systems (SIGSPATIAL '25), November 3--6, 2025,
Minneapolis, MN, USA}
\acmDOI{10.1145/3748636.3762752}
\acmISBN{979-8-4007-2086-4/2025/11}

\begin{document}

\title{Graph Enhanced Trajectory Anomaly Detection}

\author{Jonathan Kabala Mbuya}
\email{jmbuya@gmu.edu}
\orcid{0009-0004-6227-2199}
\affiliation{%
  \institution{George Mason University}
  \streetaddress{4400 University Dr}
  \city{Fairfax}
  \state{Virginia}
  \country{USA}
  \postcode{22030}
}

\author{Dieter Pfoser}
\email{dpfoser@gmu.edu}
\affiliation{%
  \institution{George Mason University}
  \streetaddress{4400 University Dr}
  \city{Fairfax}
  \state{Virginia}
  \country{USA}
  \postcode{22030}
}

\author{Antonios Anastasopoulos}
\email{antonis@gmu.edu}
\affiliation{%
  \institution{George Mason University}
  \streetaddress{4400 University Dr}
  \city{Fairfax}
  \state{Virginia}
  \country{USA}
  \postcode{22030}
}


\begin{abstract}
 Trajectory anomaly detection is essential for identifying unusual and unexpected movement patterns in applications ranging from intelligent transportation systems to urban safety and fraud prevention.
 Existing methods only consider limited aspects of the trajectory nature and its movement space by treating trajectories as sequences of sampled locations, with sampling determined by positioning technology, e.g., GPS, or by high-level abstractions such as staypoints. Trajectories are analyzed in Euclidean space, neglecting the constraints and connectivity information of the underlying movement network, e.g., road or transit networks.
 The proposed Graph Enhanced Trajectory Anomaly Detection (GETAD) framework tightly integrates road network topology, segment semantics, and historical travel patterns to model trajectory data. GETAD uses a Graph Attention Network to learn road-aware embeddings that capture both physical attributes and transition behavior, and augments these with graph-based positional encodings that reflect the spatial layout of the road network. 
 A Transformer-based decoder models sequential movement, while a multiobjective loss function combining autoregressive prediction and supervised link prediction ensures realistic and structurally coherent representations. 
 To improve the robustness of anomaly detection, we introduce Confidence Weighted Negative Log Likelihood (CW NLL), an anomaly scoring function that emphasizes high-confidence deviations. 
 Experiments on real-world and synthetic datasets demonstrate that GETAD achieves consistent improvements over existing methods, particularly in detecting subtle anomalies in road-constrained environments. These results highlight the benefits of incorporating graph structure and contextual semantics into trajectory modeling, enabling more precise and context-aware anomaly detection.
\end{abstract}

\ccsdesc[300]{Computing methodologies~Modeling and simulation}
\keywords{Anomalous Trajectories, Anomaly Detection, Graph Neural Networks, Road Network Modeling, Mobility Data}

\received{20 February 2007}
\received[revised]{12 March 2009}
\received[accepted]{5 June 2009}

\maketitle

\section{Introduction}
Trajectory anomaly detection is a crucial task in intelligent transportation systems, ride-hailing platforms, and mobility safety monitoring. The goal is to identify trajectories that significantly deviate from normal movement patterns \cite{LiuOnlineAnomalous2020, Song2018AnomalyDetectionWithRNN, DeepTEA}, which may indicate events such as detours, fraud, or irregular behavior. A wide range of approaches has been proposed, from statistical and rule-based models to deep learning and generative methods. However, most of these models fall short in road-constrained environments due to a lack of structural awareness.

Traditional approaches like iBOAT \cite{chec2013iBOAT} rely on handcrafted features and grid-based approximations that arbitrarily discretize space without regard for road layout, often mapping trajectories onto regions that do not reflect actual road connectivity. Others, such as GM-VSAE \cite{LiuOnlineAnomalous2020}, DeepTEA \cite{DeepTEA}, and LMTAD \cite{LMTAD}, model trajectories as point sequences in Euclidean space. These approaches ignore the structure of the road network, including how road segments are connected, and overlook critical contextual information. Important factors such as road type (e.g., highway vs. residential) and speed limits are not taken into account, making it difficult for these models to accurately interpret real-world movement patterns. Furthermore, CausalTAD \cite{CausalTAD} attempts to address distribution shifts through causal reasoning but compresses source-destination pairs into latent vectors without modeling the road network, limiting its ability to capture spatial relationships that distinguish anomalies from normal paths.

In this work, we introduce Graph Enhanced Trajectory Anomaly Detection (GETAD), a novel framework that directly incorporates road network structure and semantics into trajectory modeling. Unlike prior methods that define anomalies as arbitrary deviations from high-frequency paths, GETAD models trajectory anomalies as deviations from typical route behavior, where typicality is defined by structural connectivity, historical travel patterns, and road-specific characteristics such as road segment type (i.e., highway, residential), permitted direction, and usage frequency. GETAD integrates a Graph Attention Network (GAT) with a Transformer-based decoder, replacing standard token embeddings with learned embeddings of road segments that encode both local connectivity (e.g., adjacent segments) and global semantic information (e.g., road type, historical usage, directionality). This allows the model to distinguish between valid but infrequent detours and genuinely anomalous behavior that deviates from typical road-constrained travel patterns. For example, consider a driver who exits a highway to travel along a low-traffic residential road before rejoining the highway at a later point. Models that rely primarily on frequency may identify this trajectory as anomalous solely because it contains road segments that are rarely observed in the historical data. In contrast, GETAD incorporates additional contextual information when evaluating such behavior. It considers attributes such as road type, expected travel speed, historical usage patterns, and the road segment’s functional role within the overall route. By analyzing these factors jointly, GETAD is able to assess whether the detour represents a reasonable deviation, such as avoiding congestion, or a genuinely abnormal routing choice that departs from typical travel behavior.

To support anomaly detection grounded in the spatial layout of the road network, GETAD incorporates graph-based positional encodings that capture shortest-path distances between road segments, enabling the model to reason about spatial layout and topological relationships, such as unexpected detours or deviations from typical connection patterns. To jointly capture sequential dependencies and reinforce structural consistency in the learned representations, GETAD is trained using a multi-objective loss that combines autoregressive prediction, supervised link prediction, and contrastive learning. The cross-entropy loss trains the model to predict the next road segment in a trajectory based on historical context, enabling it to assign meaningful likelihoods for anomaly scoring. The link prediction loss ensures that the learned road segment embeddings preserve true transition patterns observed in the road network, reinforcing structural validity.

We also propose a new anomaly scoring function, Confidence Weighted Negative Log Likelihood (CW-NLL), which refines traditional negative log likelihood by incorporating model confidence into the score. Unlike standard scoring methods such as perplexity or raw negative log likelihood, CW-NLL accounts for the certainty with which the model assigns low probability to a prediction. This distinction is important in trajectory modeling, where multiple valid next segments may exist and low probability does not necessarily indicate abnormal behavior. CW-NLL emphasizes predictions that are both unlikely and made with high confidence, which are often the clearest indicators of true anomalies, while assigning less weight to uncertain predictions where several plausible continuations exist.
Our key contributions are as follows:

\begin{itemize}

    \item A graph-enhanced architecture that uses a Graph Attention Network (GAT) to learn road segment embeddings, capturing both spatial structure and contextual semantics within the road network.
    
    \item A multi-objective training framework that combines autoregressive next-segment prediction and supervised link prediction over road transitions. This enables the model to learn realistic movement patterns, preserve observed road connectivity, and sharpen distinctions between typical and atypical routes.
    
    \item Incorporating graph-based positional encodings derived from shortest-path distances to enrich the model’s understanding of spatial layout and topological relationships between road segments.
    
    \item A novel anomaly scoring function, Confidence Weighted Negative Log Likelihood (CW-NLL), which integrates prediction likelihood with model confidence to better distinguish between rare but plausible events and true anomalies.
    
    \item Extensive experimentation using real-world and synthetic datasets and demonstrating that GETAD outperforms existing methods in detecting both common and subtle trajectory anomalies.
\end{itemize}

The remainder of this paper is organized as follows. Section~\ref{sec:relatedwork} reviews existing approaches to trajectory anomaly detection and graph-based modeling. Section~\ref{preliminaries} introduces the formal problem definition and key concepts. In Section~\ref{sec:method}, we present the GETAD framework, including the graph-based segment embedding, positional encoding, and trajectory modeling components. Section~\ref{sec:experiments} reports on empirical evaluations using real-world and synthetic datasets. Finally, Section~\ref{sec:conclusion} concludes the paper and outlines potential directions for future research.

\section{Related Work} \label{sec:relatedwork}

\subsection{Trajectory Anomaly Detection}

Research on trajectory anomaly detection has generally followed two main directions: heuristic-based \cite{Lee2008TRAOD, Zhang2011IBat, chec2013iBOAT, Zhu2015Time_dependent_popular, Zhongjian2017} and learning-based methods \cite{Song2018AnomalyDetectionWithRNN, Smolyak2020GANsforAnomalyDection}.

Heuristic-based methods typically rely on manually designed features and rule-based similarity metrics to identify anomalous behavior. Early work such as \cite{Lee2008TRAOD} introduced a partition-and-detect framework that combines distance and density-based criteria to isolate unusual sub-trajectories. Similarly, methods like \cite{Zhang2011IBat} and \cite{chec2013iBOAT} detect anomalies by grouping trajectories according to source-destination pairs and flagging those that follow statistically rare paths. These approaches often employ isolation-based principles to separate uncommon behaviors from dominant movement patterns. Time-dependent variants, such as \cite{Zhu2015Time_dependent_popular}, adapt these comparisons to specific temporal windows, while others like \cite{Zhongjian2017} refine spatial similarity using edit distance and clustering. While these techniques are interpretable and computationally efficient, they suffer from several key limitations: their reliance on hand-crafted features makes them inflexible, their detection accuracy is sensitive to domain-specific thresholds, and they struggle to generalize across heterogeneous urban environments. In contrast, GETAD eliminates the need for manually engineered features by learning rich road segment embeddings directly from data, capturing both how segments are used and how they are connected. This allows the model to generalize across different settings and reliably separate unusual but valid routes from true anomalies.

Learning-based approaches have emerged as powerful alternatives to heuristic methods, offering the ability to automatically learn complex movement patterns from data without relying on handcrafted rules. Early work in this category leverages recurrent neural networks (RNNs) to encode trajectories into latent embeddings \cite{Song2018AnomalyDetectionWithRNN}, capturing sequential dependencies across visited locations. However, these methods typically depend on labeled data for supervised training, which is rarely available at scale due to the high cost of annotating anomalies in trajectory datasets.

To overcome this, several unsupervised or self-supervised methods have been proposed. Some rely on generative models that combine reconstruction-based objectives with learned priors. For instance, Infinite Gaussian Mixture Models coupled with bi-directional GANs have been used to detect outliers based on a mixture of reconstruction and discriminator losses \cite{Smolyak2020GANsforAnomalyDection}. Variants of autoencoders \cite{Malhotra2016LSTMbasedEF, Zhou2017AnomalyDetectionAE, Jinghui2017AEnsembles, Zong2018DeepAG} compress and reconstruct input trajectories under the assumption that anomalous behaviors lead to large reconstruction errors \cite{Zong2018DeepAG}. More recent efforts have introduced increasingly sophisticated models for unsupervised or self-supervised anomaly detection. DeepTEA~\cite{DeepTEA} captures temporal variability by modeling latent traffic patterns over time, allowing it to detect outliers that are inconsistent with typical time-dependent movement. CausalTAD~\cite{CausalTAD} uses a causal generative model to control for confounding variables that arise when source and destination pairs strongly influence the generated route. By explicitly separating these effects, it improves generalization to new or less frequent origin-destination combinations. LM-TAD~\cite{LMTAD} frames trajectory modeling as a language modeling problem, using a Transformer to predict token-level likelihoods and localize anomalies with metrics like surprisal and perplexity.

Despite their strengths, these models share important limitations. None of them incorporate the structural layout or semantics of the road network. DeepTEA focuses on temporal context, CausalTAD abstracts routes into dense latent spaces, and LM-TAD operates on token sequences without knowledge of physical connectivity or topological constraints. As a result, they may misclassify rare but valid detours or overlook subtle structural deviations.

GETAD addresses these gaps by explicitly modeling road network structure using a Graph Attention Network (GAT), combined with a Transformer-based decoder. It learns segment embeddings that encode connectivity and semantic features, enabling reasoning about both sequential behavior and structural consistency.

\subsection{Graph Learning on Trajectory Data}

Graph-based learning has been increasingly applied to trajectory representation, motivated by the observation that movement patterns are constrained by the connectivity and semantics of the road network \cite{su17083705, START_2024, Zhang2025}. Several works have used graph neural networks (GNNs) or attention-based variants such as GATs to learn the embeddings of road segments~\cite{toast_2021, START_2024}. Toast learns road segment embeddings by capturing both traffic context and local route consistency, enabling general-purpose spatial representations~\cite{toast_2021}. START extends this by incorporating temporal regularities and travel semantics, such as typical speeds and visit durations, into a GAT-based encoder for better trajectory modeling~\cite{START_2024}. These models incorporate topological features (e.g., node degrees, transition probabilities) and road attributes (e.g., type, speed limit), enabling semantically rich representations useful for downstream tasks.

However, the majority of such graph-enhanced models have been developed for trajectory prediction or trajectory similarity~\cite{toast_2021}, not anomaly detection. They are typically trained to forecast the most likely next segment given current and past segments, and do not explicitly model or evaluate the likelihood of deviation from normal behavior. As such, they are not equipped to flag distributional outliers or structural violations in observed trajectories.

Furthermore, in many of these models, graph-based representations are learned independently from the sequence model, either precomputed or used statically. This decoupling limits their ability to reflect dynamic contextual influences on route choice, such as time of day, behavioral patterns, or road conditions.

GETAD departs from this separation by tightly integrating graph-based representations into a generative sequence model. Segment embeddings learned by the GAT are updated jointly with the Transformer decoder, aligning structural knowledge with sequential behavior.

\section{Preliminaries}
\label{preliminaries}

Important aspects of this work include the foundational concepts of road networks, raw trajectories, map-matched trajectories, and the formulation of the online trajectory anomaly detection problem.

\subsection{Road Network}

We represent the road network as a directed graph $\mathcal{G} = (\mathcal{V}, \mathcal{E})$, where $\mathcal{V}$ denotes the set of nodes and $\mathcal{E} \subseteq \mathcal{V} \times \mathcal{V}$ denotes the set of directed edges. Each node $v \in \mathcal{V}$ corresponds to a distinct road segment with a specific direction of travel. An edge $(v_i, v_j) \in \mathcal{E}$ denotes a feasible transition from segment $v_i$ to segment $v_j$ based on physical connectivity on a road network.

Each node in the graph is associated with descriptive attributes such as segment length, road type, average traffic volume, or historical usage frequency. These attributes are encoded in a feature matrix $\mathbf{F} \in \mathbb{R}^{|\mathcal{V}| \times d_0}$, where $|\mathcal{V}|$ is the number of segments, and $d_0$ is the number of input features. The connectivity between segments is captured by the adjacency matrix $\mathbf{A} \in \{0, 1\}^{|\mathcal{V}| \times |\mathcal{V}|}$, where $\mathbf{A}_{ij} = 1$ indicates a valid transition from $v_i$ to $v_j$. 

\subsection{Map-Matched Trajectory}

A raw trajectory records the path of a moving object over time, typically captured through GPS-enabled devices. Formally, a raw trajectory is defined as an ordered sequence of spatial-temporal points:
\[
T^{raw} = \langle p_1, p_2, \dots, p_n \rangle,
\]
where each point $p_i = (x_i, y_i, t_i)$ consists of geographic coordinates $(x_i, y_i)$ and a timestamp $t_i$. Though these trajectories provide rich behavioral data, they lack explicit alignment with the underlying road structure as they often exhibit irregular sampling intervals, missing data, and location noise. To address the limitations of raw GPS trajectories, a map-matching algorithm is used to align each trajectory with the underlying road network $\mathcal{G}$. In essence, map-matching involves projecting each measured GPS point to the most likely location on a road segment, thereby correcting for noise and ensuring that the resulting trajectory follows feasible paths in the road network. A map-matched trajectory is thus defined as a chronological sequence of road segments:
\[
T = \langle v_1, v_2, \dots, v_m \rangle, \quad v_i \in \mathcal{V},
\]
where each $v_i$ denotes a road segment that the object is estimated to have traversed. This representation abstracts away GPS noise, aligns trajectories with the road network topology, and facilitates the extraction of structured features. 

\subsection{Problem Formulation}

Given a road network represented as a directed graph $\mathcal{G} = (\mathcal{V}, \mathcal{E})$, and a collection of historical map-matched trajectories $\mathcal{T}_{\text{train}} = \{ T_1, T_2, \dots, T_N \}$, where each trajectory $T_i = \langle v_1, v_2, \dots, v_m \rangle$ is a sequence of road segments with $v_j \in \mathcal{V}$, the objective is to detect anomalous behavior in newly observed trajectories (or sub-trajectories) in an online manner. 

More specifically, given a test trajectory $T_{\text{test}}$, the task is to assign an anomaly score that indicates whether the trajectory deviates from the normal patterns inferred from $\mathcal{T}_{\text{train}}$. Anomalous behavior may include irregular detours, deviations from frequently traveled routes, or patterns that are inconsistent with the spatial or temporal characteristics of normal trajectories. In this paper, we specifically focus on detour anomalies, operationalized as deviations in the spatial path along the road network. While irregular behaviors such as erratic driving patterns or fraud detection are important directions, our experimental scope is restricted to detours for reproducibility and fair comparison with existing baselines.
\section{Method} \label{sec:method}

\begin{figure}
\includegraphics[width=1 \columnwidth]{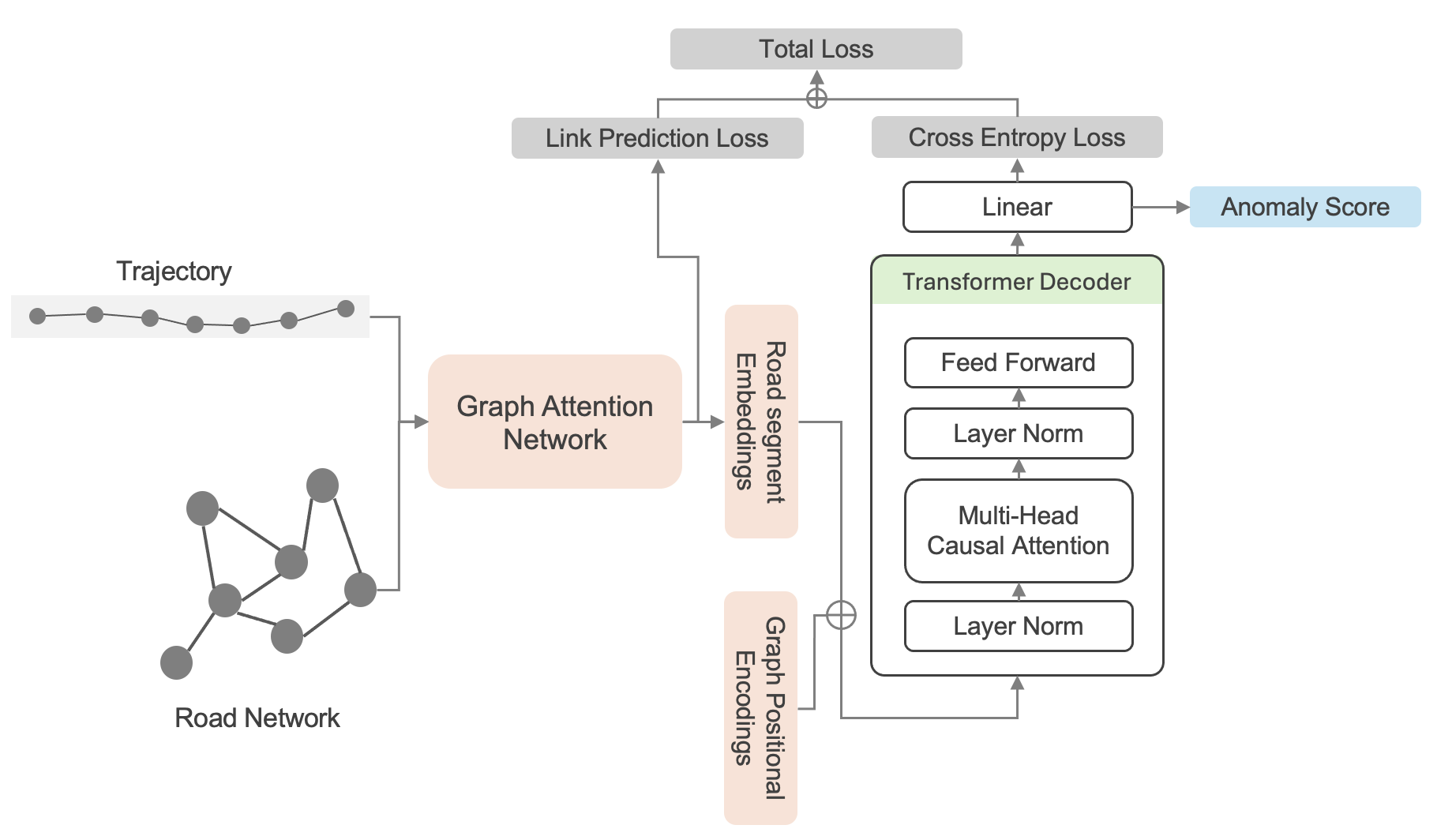}
\caption{GETAD architecture. }
\label{fig:model}
\Description{}
\end{figure}  

The proposed approach comprises the following main components and steps. 
We first present a method for \textit{learning semantically rich representations of road segments using a Graph Attention Network}, which incorporates both physical attributes and observed transition behavior. 
Second, a Graph Positional Embedding module encodes the spatial layout through shortest-path distances in the road network. These two components are combined within a Transformer Decoder to model trajectory sequences in an autoregressive manner. 
A multi-objective learning framework is used to train the model, including cross-entropy and link prediction losses. Finally, a confidence-weighted likelihood-based anomaly score is designed to emphasize high-certainty deviations, and we discuss how this formulation improves robustness in detecting irregular movement patterns. 
Figure \ref{fig:model} shows the overall architecture of our framework.
 
\subsection{Graph-Based Road Segment Embedding}

Road networks inherently possess complex structural relationships and dynamic usage patterns, which are crucial for accurately modeling mobility behaviors and detecting anomalies. Traditional embedding methods may overlook critical nuances such as varying road segment importance and transition frequencies, resulting in less informative representations. To address these limitations, we employ a Graph Attention Network (GAT) \cite{GAT} to generate semantically rich and contextually aware embeddings for each road segment \( v \in \mathcal{V} \). The GAT architecture is particularly suited for this task because it adaptively aggregates information from neighboring segments by weighting their contributions based on local structural features and historically observed movement patterns. This design ensures that each embedding effectively encodes both the physical attributes and real-world utilization of road segments, providing a comprehensive foundation for downstream trajectory modeling and anomaly detection tasks.

Specifically, to create these informative segment embeddings, we first initialize each road segment with a feature vector \( \mathbf{h}^0_i \in \mathbb{R}^{d_0} \) constructed by concatenating several attributes. These include static infrastructure properties such as segment length, categorical road type, number of lanes, in-degree, and out-degree, as well as dynamic statistics such as the normalized visit frequency computed from observed trajectories. This design embeds both physical and behavioral characteristics of the road network.

To model relationships between connected segments, we use a multi-head GAT. For each segment \( v_i \) and its neighbors \( v_j \in \mathcal{N}(v_i) \), the attention weight is informed not only by the features of \( v_i \) and \( v_j \), but also by the empirical transition probability \( p_{v_i v_j} \), computed as:
\[
p_{v_i v_j} = \frac{\text{count}(v_i \rightarrow v_j)}{\text{count}(v_i)},
\]
where \( \text{count}(v_i \rightarrow v_j) \) is the number of times the segment pair \( (v_i, v_j) \) appears consecutively in the trajectory data.

For each attention head \( k = 1, \dots, H_1 \) in a given layer $l = 1, \ldots, L$, the unnormalized attention score is computed as:
\[
e_{v_i v_j}^{l,k} = (\mathbf{a}_1^{l,k})^\top \cdot \text{LeakyReLU} \left( \mathbf{W}_1^{l,k} \mathbf{h}_i^l + \mathbf{W}_2^{l,k} \mathbf{h}_j^l + p_{v_i v_j} \cdot \mathbf{a}_2^{l,k} \right),
\]
where \( \mathbf{W}_1^{l,k}, \mathbf{W}_2^{l,k} \in \mathbb{R}^{d_l \times d_{l+1}} \) are learned projection matrices, and \( \mathbf{a}_1^{l,k}, \mathbf{a}_2^{l,k} \in \mathbb{R}^{d_l} \) are attention parameter vectors. The scalar transition probability modulates the attention signal, emphasizing frequently observed transitions in historical data.

These scores are normalized across neighbors using the softmax function:
\[
\alpha_{v_i v_j}^{l,k} = \frac{\exp(e_{v_i v_j}^{l,k})}{\sum_{n \in \mathcal{N}(v_i)} \exp(e_{v_i n}^{l,k})}.
\]

The output of each attention head is computed by a weighted sum over the transformed neighbor features:
\[
\mathbf{h}_i^{l+1,k} = \sum_{j \in \mathcal{N}(v_i)} \alpha_{v_i v_j}^{l,k} \mathbf{W}_3^{l,k} \mathbf{h}_j^l,
\]
where \( \mathbf{W}_3^{l,k} \in \mathbb{R}^{d_l \times d_{l+1}} \) is the output projection matrix for head \( k \).

The final embedding for node \( v_i \) is obtained by concatenating the outputs from all heads and applying a non-linear activation function \( \sigma \) (e.g., ELU \cite{elu_2015} or ReLU \cite{relu_2010}):
\[
\mathbf{h}_i = \sigma\left( \big\|_{k=1}^{H} \mathbf{h}_i^{l+1,k} \right).
\]

This attention mechanism enables the model to selectively aggregate information from neighboring road segments based on their structural similarity, feature attributes, and observed co-occurrence frequency. The resulting segment embeddings provide a semantically and spatially informed representation of the road network, which is passed to the downstream trajectory encoder.

\subsection{Graph Positional Embedding}

To enrich trajectory representations with spatial context, we introduce a Graph Positional Embedding (GPE) mechanism that encodes the relative positions of road segments based on their connectivity within the road network graph. Unlike traditional positional encodings that rely on sequential indices, GPE captures structural relationships among segments using hop distances, allowing the model to reason about spatial layout and graph topology. This design is inspired by recent work on relative and graph-aware positional encodings \cite{shaw2018self, dwivedi2021graph, you2021graphbert}, but adapted specifically for road-constrained trajectories.

Given a trajectory composed of \( |T| \) road segments, we construct a \( |T| \times |T| \) distance matrix \( \mathbf{D} \), where each entry \( \mathbf{D}_{ij} \) denotes the shortest-path hop distance between segments \( v_i \) and \( v_j \) in the road network graph \( \mathcal{G} \). If a segment pair is not connected, we assign a fixed maximum distance value \( d_{\max} \). Each distance is then mapped to a learnable embedding using a lookup table:
\[
\mathbf{P}_{ij} = \mathbf{E}_{\text{dist}}[\mathbf{D}_{ij}], \quad \mathbf{E}_{\text{dist}} \in \mathbb{R}^{(d_{\max} + 1) \times d_{\text{model}}},
\]
where \( d_{\text{model}} \) is the dimensionality of the positional embeddings.

To account for special tokens in the trajectory, such as start-of-trajectory (SOT), end-of-trajectory (EOT), and padding (PAD) tokens, we define dedicated learnable embeddings \( \mathbf{e}_{\text{sot}}, \mathbf{e}_{\text{eot}}, \mathbf{e}_{\text{pad}} \in \mathbb{R}^{d_{\text{model}}} \).

We then compute the final positional embedding for each token at index \( i \) by aggregating the distance-based embeddings relative to all other tokens in the trajectory:
\[
\mathbf{ge}_i = \sum_{j=1}^{|T|} \mathbf{P}_{ij}
\]

This graph-aware positional encoding captures both local proximity and global structural context within the trajectory. By encoding how each road segment is situated relative to others in the network, GPE enhances the model's ability to distinguish between normal sequences and subtle spatial deviations. These positional embeddings are added to the input token embeddings and passed to the Transformer decoder.

\subsection{Trajectory Modeling with Transformer Decoder}

After constructing semantically rich road segment representations using the Graph Attention Network (GAT) and embedding topological relationships through the Graph Positional Embedding (GPE) module, we model the sequential nature of trajectories using a Transformer Decoder. This component captures temporal dependencies, order-sensitive correlations, and contextual patterns that evolve across the trajectory sequence.

Each trajectory \( T = \langle v_1, v_2, \dots, v_m \rangle \) is represented as a sequence of road segments, where each segment \( v_i \) is embedded as:
\[
\mathbf{z}_i = \mathbf{h}_{v_i} + \mathbf{ge}_{v_i},
\]
with \( \mathbf{h}_{v_i} \) denoting the road segment embedding produced by the GAT, and \( \mathbf{ge}_{v_i} \) the positional vector derived from the GPE module. The resulting sequence \( \mathbf{Z} = \langle \mathbf{z}_1, \dots, \mathbf{z}_m \rangle \) forms the input to the Transformer Decoder \cite{Vaswani2017AttentionIA}.

To preserve the causal structure of trajectories, we adopt an autoregressive decoding setup, where the model learns to predict the next road segment based only on previously observed segments. The decoder consists of \( M \) stacked layers, each containing a masked multihead self-attention mechanism followed by a positionwise feedforward network. Causal masking ensures that, for any position \( i \), the model only attends to tokens at positions \( j \leq i \), thereby maintaining the temporal progression of the sequence.

Within each layer, multi-head attention is used to capture a range of relational patterns between positions. For each attention head \( h = 1, \dots, H_2 \), the input embeddings are projected into queries, keys, and values:
\[
\mathbf{Q}^{(h)} = \mathbf{Z} \mathbf{W}_Q^{(h)}, \quad 
\mathbf{K}^{(h)} = \mathbf{Z} \mathbf{W}_K^{(h}, \quad 
\mathbf{V}^{(h)} = \mathbf{Z} \mathbf{W}_V^{(h},
\]
where \( \mathbf{W}_Q^{(h)}, \mathbf{W}_K^{(h)}, \mathbf{W}_V^{(h)} \in \mathbb{R}^{d \times d_h} \) are learned projection matrices and \( d_h = d / H_2 \) is the dimension of each attention head.

The output of the scaled dot product attention for head \( h \) is given by:
\[
\text{Attention}^{(h)}(\mathbf{Q}, \mathbf{K}, \mathbf{V}) = \text{softmax} \left( \frac{\mathbf{Q}^{(h)} (\mathbf{K}^{(h)})^\top}{\sqrt{d_h}} + \mathbf{M} \right) \mathbf{V}^{(h)},
\]
where \( \mathbf{M} \in \mathbb{R}^{m \times m} \) is the causal mask that assigns a large negative value to all future positions.

The outputs of all attention heads are concatenated and projected through a linear transformation:
\[
\text{MultiHead}(\mathbf{Z}) = \text{Concat}\left( \text{Attention}^{(1)}, \dots, \text{Attention}^{(H_2)} \right) \mathbf{W}_O,
\]
where \( \mathbf{W}_O \in \mathbb{R}^{d \times d} \) is a learned projection matrix.

This output is passed through a residual connection, followed by layer normalization \cite{ba2016layernormalization}  and a positionwise feedforward network:
\[
\text{FFN}(x) = \text{ReLU}(x \mathbf{W}_1 + \mathbf{b}_1) \mathbf{W}_2 + \mathbf{b}_2.
\]

Each decoder layer applies this sequence of operations, producing a progressively refined embedding for each token. The final output of the Transformer Decoder at each position is a contextualized representation that integrates spatial structure from the road network, positional context from the GPE, and sequential dynamics learned through autoregressive modeling.

\subsection{Learning Objectives}

The model is trained using a multi-objective loss that combines two complementary signals: cross-entropy loss for autoregressive sequence modeling and link prediction loss \cite{zhang2018linkpredictionbasedgraph, hisano2016semisupervisedgraphembeddingapproach} to enforce structural consistency in the road network. Together, these objectives enhance the model’s ability to capture both sequential movement patterns and underlying graph connectivity, enabling better generalization to diverse trajectory behaviors.

\paragraph{Cross-Entropy Loss ($\mathcal{L}_{\text{CE}}$):}
to model the probability distribution over trajectory sequences, we use a standard cross-entropy loss applied to the output of the Transformer decoder. Given a map-matched trajectory $T = \langle v_1, v_2, \dots, v_m \rangle$, the model learns to predict the next segment $v_i$ conditioned on its historical context $\langle v_1, \dots, v_{i-1} \rangle$. The objective is to minimize:

\[
\mathcal{L}_{\text{CE}} = - \sum_{i=1}^{m} \log P(v_i \mid v_1, \dots, v_{i-1}; \theta).
\]

This loss captures the sequential dependencies of trajectory patterns and enables the model to assign likelihoods to full sequences or sub-trajectories, which is crucial for anomaly detection.

\paragraph{\textbf{Link Prediction Loss ($\mathcal{L}_{\text{link}}$):}}
to ensure that the learned graph-based embeddings reflect the true structure of the road network, we include a supervised link prediction objective. Given a pair of nodes $(u, v)$ and a binary label $y_{uv} \in \{0, 1\}$ indicating whether a transition between them exists in the historical data, the model is trained to distinguish true transitions from random pairs. The loss is defined as:

\[
\mathcal{L}_{\text{link}} = - \sum_{i=1}^{N} \left[ y_i \log \sigma(s_i) + (1 - y_i) \log (1 - \sigma(s_i)) \right],
\]

where $s_i = \langle \mathbf{z}_u, \mathbf{z}_v \rangle$ is the similarity score between road segment embeddings and $\sigma(\cdot)$ is the sigmoid function. This objective encourages the GAT encoder to produce embeddings that preserve topological and transition semantics in the road network.

\paragraph{\textbf{Final Objective:}}
the full training objective is a weighted combination of the three loss components:

\[
\mathcal{L}_{\text{total}} = \lambda_{\text{CE}} \mathcal{L}_{\text{CE}} + \lambda_{\text{link}} \mathcal{L}_{\text{link}}
\]

where $\lambda_{\text{CE}}, \lambda_{\text{link}}$ are tunable hyperparameters controlling the contribution of each objective.

Together, these objectives guide the model towards learning meaningful spatial embeddings, capturing movement patterns, and distinguishing between normal and anomalous behaviors in a unified training framework.

\subsection{Anomaly Score}

A novel anomaly scoring function, Confidence-Weighted Negative Log-Likelihood (CW-NLL), refines the traditional negative log-likelihood (NLL) by incorporating model confidence into the anomaly score. Standard anomaly scoring techniques, such as perplexity or raw NLL, assess how unlikely a token is under the model’s learned distribution, but do not distinguish between low-likelihood predictions made with high certainty and those made with uncertainty. This limitation becomes critical in trajectory modeling, where the model may assign low probability to a token simply because the context allows for multiple valid continuations. In such cases, the uncertainty reflects natural ambiguity rather than true anomaly, and treating all low-likelihood events as anomalous can lead to false positives. 

CW-NLL addresses this by weighting the contribution of each token's NLL according to the model’s confidence, allowing the score to focus on predictions that are both unlikely and made with high certainty.

This approach draws on the observation that neural networks often produce high-confidence predictions even when the input is out-of-distribution or anomalous~\cite{guo2017calibration,nguyen2015deep,hein2019relu}. Unlike methods that rely solely on the maximum softmax probability as a proxy for confidence~\cite{hendrycks2017baseline}, CW-NLL leverages the entropy of the predictive distribution at each step to weight token-level likelihoods. This does not represent global model certainty but instead reflects the distributional sharpness of predictions within a trajectory context. By combining negative log-likelihood with entropy-based weights, CW-NLL emphasizes low-likelihood predictions made with high distributional confidence and reduces the influence of uncertain predictions where multiple outcomes are plausible.

Let $\ell_i$ be the cross-entropy loss at position $i$, and let $\mathbf{p}_i$ be the softmax distribution over the vocabulary of road segments at that timestep.

The entropy of this distribution is given by:
\[
H(\mathbf{p}_i) = -\sum_j \mathbf{p}_i(j) \log \mathbf{p}_i(j)
\]

The maximum possible entropy is $H_{\text{max}} = \log V$ where $V$ is the vocabulary size. The confidence is then defined as:
\[
c_i = 1 - \frac{H(\mathbf{p}_i)}{H_{\text{max}}}
\]

with values close to 1 indicating high certainty. The final CW-NLL score for a trajectory of length $n$ is computed as:
\[
\text{CW-NLL} = \sum_{i=1}^{n} c_i \cdot \ell_i
\]

This formulation emphasizes low-likelihood predictions made with high confidence, which are the most indicative of true anomalies, while assigning less weight to uncertain predictions where multiple outcomes may be plausible. As a result, CW-NLL captures not only the unexpectedness of a token but also the degree of certainty with which the model assessed its likelihood, producing a more focused and reliable anomaly signal in trajectory data.

\section{Experiments} \label{sec:experiments}
\subsection{Datasets \& Preprocessing}

Our experimentation leverages synthetic (agent-based models) and real-world datasets. Specifically, we use a synthetic Pattern of Life dataset (cf. \cite{ZuflePatternOfLife2023}) and the Porto taxi dataset \cite{YuanTdriveDrivingPorto, YuanDrivingWithKnolwedgePorto}.

\begin{table}[h]
\caption{Statistics of trajectory dataset}
\label{tab:dataset_stats}
\centering
\begin{tabular}{rccc} \toprule
                   

Dataset & \# Points &  \# Trajectories & \# Road segments \\ \midrule
 
Simulation & 68,454,977 & 1,049,458 & 27,096  \\ \midrule
Porto & 33,462,225 & 852,439 & 10,533 \\ \bottomrule
                 
\end{tabular}
\end{table}

\subsubsection{Synthetic Dataset} 
\label{sim_dataset_description}
The synthetic dataset was generated using an Agent-Based Model leveraging Patterns-of-Life  \cite{ZuflePatternOfLife2023, amiri2023massive}. This simulation consists of virtual agents designed to emulate humans' needs and behavior by performing activities in the city of San Francisco. Activities include going to work, restaurants, visiting recreational places and running errands. The trips occur on a real road network derived from the respective OpenStreetMap data \cite{OpenStreetMap2023}. The agent movement is recorded as trajectories. In addition, we also capture the staypoints (i.e., home, work, restaurant) and their respective timestamps. This allows us to experiment with different trajectory representations. 
Unlike existing approaches \cite{ZuflePatternOfLife2023, amiri2023massive} that rely only on staypoints, this dataset contains the complete trajectories connecting staypoints. Our dataset captures the movement of 5000 agents during a two-month period. Table~\ref{tab:dataset_stats} provides a numerical summary of our data. 
We filtered trajectories with fewer than 25 road segments and considered only agents with more than 100 trajectories. 
Given that they are the output of a simulation that leverages a road network, the trajectories in this dataset are already map-matched. Our models use a sequence of road segment IDs as input.


\begin{figure*}
    \centering
    \begin{subfigure}[b]{0.4\textwidth}
        \includegraphics[width=\textwidth]{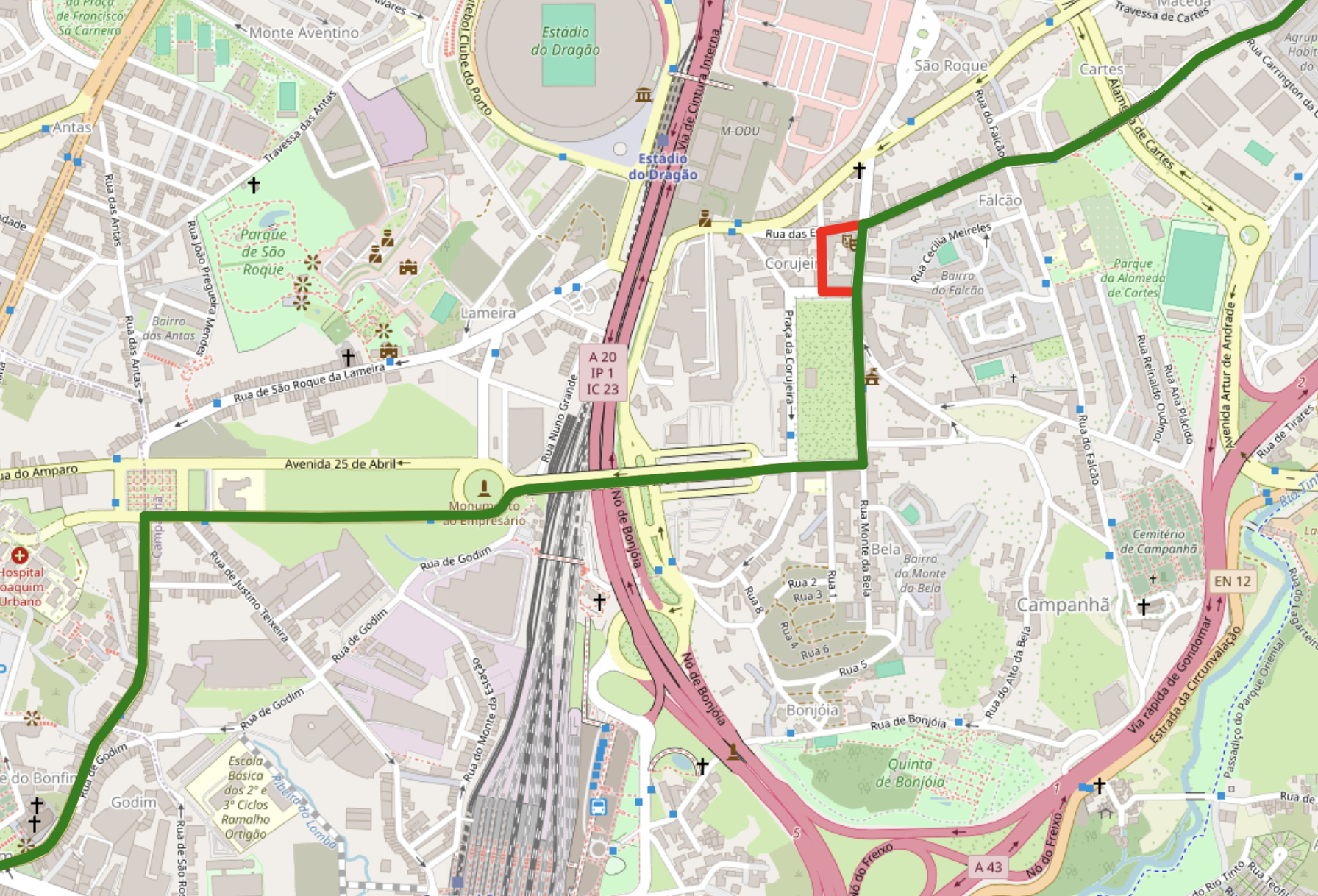}
        \caption{Constrained anomalies}
        \label{fig:random_shift_outlier}
    \end{subfigure}
    \hspace{0.05\textwidth} 
    \begin{subfigure}[b]{0.25\textwidth}
        \includegraphics[width=\textwidth]{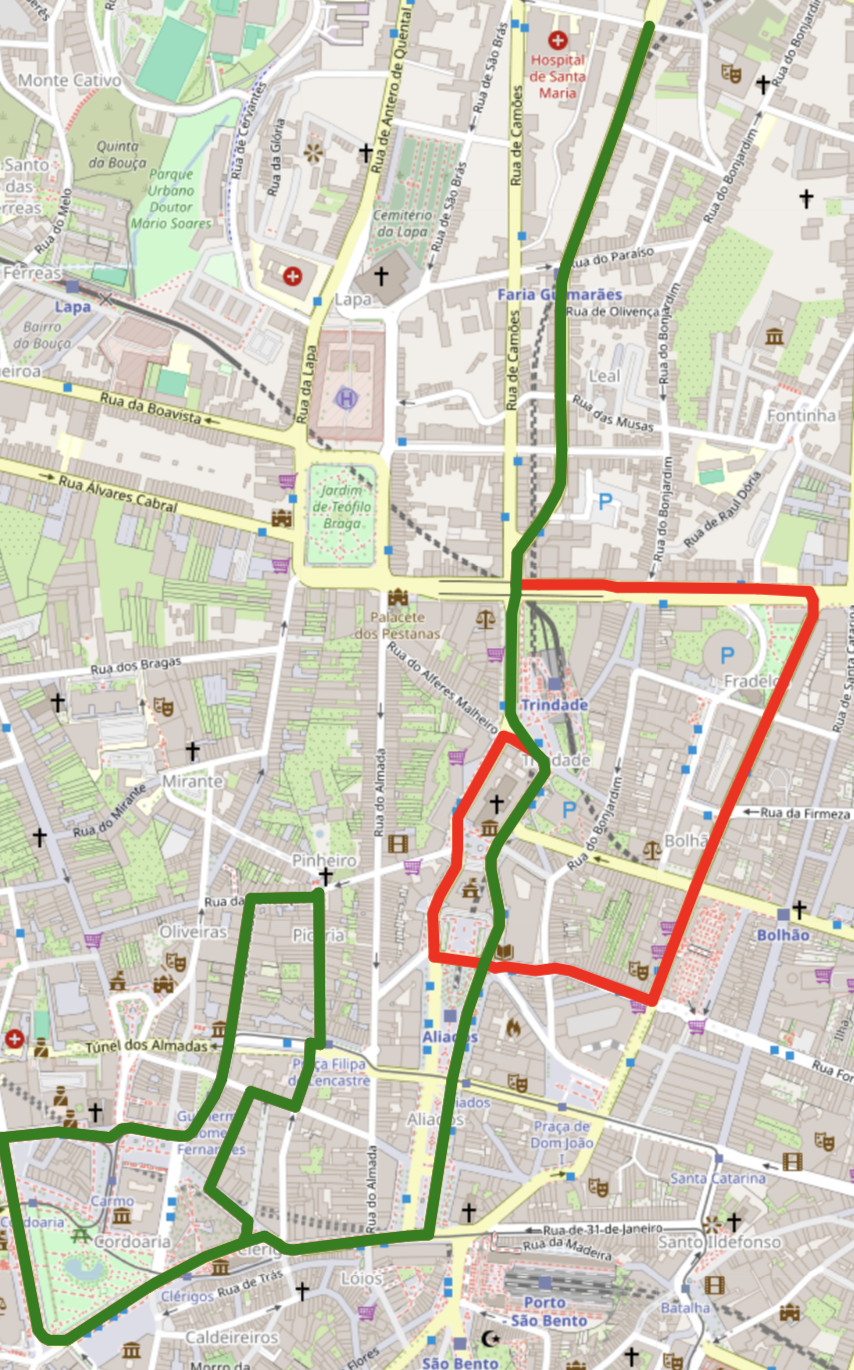}
        \caption{Unconstrained anomalies}
        \label{fig:detour_outlier}
    \end{subfigure}
    \caption{Example of generated anomalies on the Porto dataset. The detour is represented by the red trajectory. The unconstrained anomalies tend to be longer than the constrained ones, as the detour length is not limited in the unconstrained setting.}
    \label{fig:porto_outliers}
\end{figure*}

\subsubsection{Porto Dataset}
The Porto dataset \cite{moreira2013taxi} consists of data generated by 442 taxis operating in the city of Porto in Portugal (Jan 07, 2013 - June 30, 2014). A taxi reports its GPS location at 15s intervals. Following the pre-processing steps discussed in \cite{START_2024}, the raw trajectories are map-matching to the road network using the Fast Map Match algorithm \cite{fmm2018} to obtain road-contained trajectories. We use a road network derived from 
OpenStreetMap (OSM) \cite{OpenStreetMap2023} to construct the directed graph (road network) $\mathcal{G} = (\mathcal{V}, \mathcal{E})$ alongside the adjacency matrix $\mathbf{A}$ and feature matrix $\mathbf{F}$ as discussed in Section \ref{preliminaries}. Road type, length, number of lanes, maximum travel speed, the in-degree and out-degree are used to construct the feature matrix $\mathbf{F}$. Table \ref{tab:dataset_stats} provides a numerical summary of this dataset.

\subsubsection{Ground Truth}
As there are no publicly available labeled datasets for trajectory anomaly detection, two main options are available for constructing ground truth data, (i) manual annotation or (ii) synthetic anomaly generation. With manual labeling being labor intensive, prone to inaccuracies, and hard to scale, we adopted an anomaly generation strategy inspired by prior work \cite{LiuOnlineAnomalous2020, DeepTEA, CausalTAD}, to provide for a scalable and reliable evaluation.

Our method is specifically designed for road networks and respects its constraints, unlike previous efforts that use a grid cell to
create anomalies that may violate road network constraints (i.e., a road that crosses a river without a bridge). Given a trajectory $T = \langle v_1, v_2, \dots, v_m \rangle, \quad v_i \in \mathcal{V}$, we create \textit{a detour anomaly}. A detour trajectory is created by temporarily removing one or more contiguous road segments, $v_i$ to $v_j$, from the network and recalculating a new path between $v_{i-1}$ and $v_{j+1}$ using a shortest path algorithm. The resulting sub-path replaces the original segment to simulate a realistic detour. In this work, we explore two variants of the detour anomaly: the constrained and the unconstrained. 
The unconstrained variant does not constrain the length of the detour. The constrained version limits the size of the detour to be no longer than 20\% of the original trajectory. Figure~\ref{fig:porto_outliers} shows an example of the two types of anomalies.

Using these strategies, we create two separate anomalous test sets: one for constrained anomalies and another for unconstrained anomalies. Each dataset is constructed independently by selecting 5 percent of the normal trajectories and converting them into anomalies using the corresponding detour strategy. These anomalous trajectories are then combined with the remaining normal trajectories to form two distinct evaluation sets, each reflecting a different type of anomaly.

We note that while these simulated detours cannot capture all the complexities of real-world deviations (e.g., irregular but small local detours or behavioral anomalies like erratic driving), they provide a controlled and scalable benchmark. By constructing both constrained and unconstrained detours directly over the road network, we ensure that generated anomalies remain structurally valid and reproducible while still capturing a wide range of deviation magnitudes.

\subsubsection{Tokenization \& Vocabulary}

To apply sequence modeling techniques to trajectory data, all methods in our evaluation require a discrete vocabulary over which sequences are defined. In our case, we treat each unique road segment as a vocabulary token, allowing trajectories to be modeled as sequences of road segment identifiers (tokens).

For each dataset, we construct the vocabulary by collecting all road segments that appear in the training trajectories. Segments not observed during training are excluded to ensure consistency across models. To support autoregressive modeling and batch processing, we also include three special tokens in the vocabulary: \textbf{SOT} (start of trajectory), \textbf{EOT} (end of trajectory), and \textbf{PAD} (used to align sequences to a fixed length during batch training).

This approach provides a consistent and compact representation of trajectories that can be used across all models in a comparable way.

\subsection{Baselines} 
To demonstrate the performance advantage of GETAD, we compare it to the following existing unsupervised trajectory anomaly detection approaches.


\textbf{SAE} \cite{Malhotra2016LSTMbasedEF, An2015VariationalAB} encodes trajectory sequences into latent representations and models normal behavior through reconstruction, using the reconstruction error to detect anomalies.

\textbf{VSAE}~\cite{An2015VariationalAB, Sachdeva2018SequentialVA} is a variational autoencoder-based approach that models trajectory sequences within a latent probabilistic space, capturing the variability of normal behavior through sequence-to-sequence encoding and decoding. Anomalies are detected based on the reconstruction error.

\textbf{GM-VSAE} \cite{LiuOnlineAnomalous2020} extends VSAE by introducing a Gaussian Mixture prior in the latent space, modeling different normal route types and anomaly detection based on the reconstruction error.


\textbf{LM-TAD} \cite{LMTAD} models trajectories as sequences using an autoregressive Transformer with causal attention, learning the probability distribution over trajectories and detecting anomalies by identifying low-probability locations based on the learned model.

\subsection{Evaluation Metrics}
We use Precision-Recall AUC and F1 scores to evaluate the performance of our method and the baseline methods. These metrics are suitable for assessing the performance of anomaly detection methods, as the number of anomalies in each dataset is small compared to normal trajectories \cite{LiuOnlineAnomalous2020, Zhang2023ATO}. For the Porto and the Simulation datasets, these metrics are computed across all trajectories.

\begin{table*}[t]
\caption{Anomaly detection results on the Porto dataset. The best results for a particular metric in a specific category are \textbf{bolded} and the second-best results are underlined.}
\label{tab:porto_results}
\centering
\begin{tabular}{r|cc|cc|cc|cc} \toprule
                   
& \multicolumn{4}{c|}{Simulation}    &  \multicolumn{4}{c}{Porto} \\ 

anomalies type:  &   \multicolumn{2}{c|}{constrained} &   \multicolumn{2}{c|}{unconstrained} &   \multicolumn{2}{c|}{constrained} &   \multicolumn{2}{c}{unconstrained} \\ \midrule
 
Metric & F1 & PR-AUC & F1 & PR-AUC & F1 & PR-AUC & F1 & PR-AUC  \\ \midrule
\multicolumn{2}{l}{\ baselines:} \\[-.5em]

SAE & 0.836 & 0.72 & 0.843 & 0.711 &  0.575 &  0.566 &  0.590 &  0.598  \\ 
VSAE & 0.849 & 0.696 & 0.833  &  0.703  &  0.579 &  0.564 &  0.593 & 0.609    \\ 
GM-VSAE & 0.772 & 0.839  & 0.711  &  0.777 &0.677& 0.664 & \textbf{0.769} &  \underline{0.796}    \\
LM-TAD & \underline{0.934}  & \underline{0.973} & \underline{0.922}  & \underline{0.965} & \underline{0.695}  & \underline{0.732} & 0.744  & 0.794  \\ \midrule
GETAD & \textbf{0.955} & \textbf{0.988}  & \textbf{0.950} & \textbf{0.983} & \textbf{0.701} & \textbf{0.756} & \underline{0.756}  &   \textbf{0.825}   \\ \bottomrule
                 
\end{tabular}
\end{table*}

\section{Experimental Evaluation Results}


The anomaly detection results of GETAD and the four unsupervised baselines for the Simulation and Porto datasets is shown in Table~\ref{tab:porto_results}. The results capture constrained and unconstrained anomaly settings. Results are reported using Precision–Recall AUC (PR-AUC) and F1 score, two widely used metrics for imbalanced anomaly detection tasks \cite{LiuOnlineAnomalous2020, Zhang2023ATO}.

\subsection{Simulation Data Performance}

Under both constrained and unconstrained anomaly settings, GETAD achieves the highest performance across all metrics, with F1 scores of 0.955 and 0.95, and PR AUC values of 0.988 and 0.983, respectively. LM-TAD follows closely with the second-best results. These two models are distinguished by their autoregressive formulation, which conditions each prediction on the specific sequence of prior locations for a given agent. This enables them to learn detailed temporal patterns and accurately estimate the likelihood of each next step in a trajectory. For example, if an agent typically travels along a familiar route but suddenly takes a different path that includes road segments they have never used before, those unfamiliar segments receive low probability under the model, flagging the sequence as anomalous. A key strength of these autoregressive models lies in their ability to maintain and attend to the agent’s identity throughout the sequence using the attention mechanism. This allows them to capture long-range dependencies and behavioral context that are unique to each agent, making them particularly effective at detecting subtle deviations in individual behavior over time, as reflected in their strong performance on this dataset.

Another important observation is that performance does not significantly differ between constrained and unconstrained anomalies in this dataset. This can be attributed to the spatial structure of the San Francisco road network, which forms the basis of the Simulation data. The network is dense, well connected, and offers many short and efficient alternative routes between any two locations. Because of this, even when anomalies are unconstrained and allowed to deviate more freely from the original trajectory, the resulting paths do not differ substantially from those generated under the constrained setting. In many cases, the unconstrained detours overlap with or closely resemble constrained ones in both distance and layout. This overlap reduces the practical difference between the two anomaly types, leading to comparable detection difficulty across both settings.

Interestingly, both SAE and VSAE outperform GM-VSAE in terms of F1 score under the constrained anomaly setting and the Simulation dataset, even though GM-VSAE is designed to be a more expressive model through the use of a Gaussian mixture prior. This result is somewhat unexpected given the strong regularities in the Simulation data, where agents follow consistent behavioral patterns over time. One explanation is that the increased expressivity of GM-VSAE makes it more prone to overfitting, capturing fine-grained variations in the training data that do not always correspond to meaningful deviations at test time. This can reduce its effectiveness when using a fixed threshold for classification. In contrast, SAE and VSAE, which rely on simpler latent spaces, tend to produce a sharper separation between normal and anomalous trajectories under a specific decision boundary, resulting in higher F1 scores. However, when performance is evaluated across a range of thresholds using PR-AUC, GM-VSAE consistently ranks higher than the other two. This indicates that while GM-VSAE may not achieve the best F1 score at a single threshold, it performs better when the evaluation considers the full range of decision boundaries.

\begin{figure*}[t]
  \centering
  \includegraphics[width=0.8\textwidth]{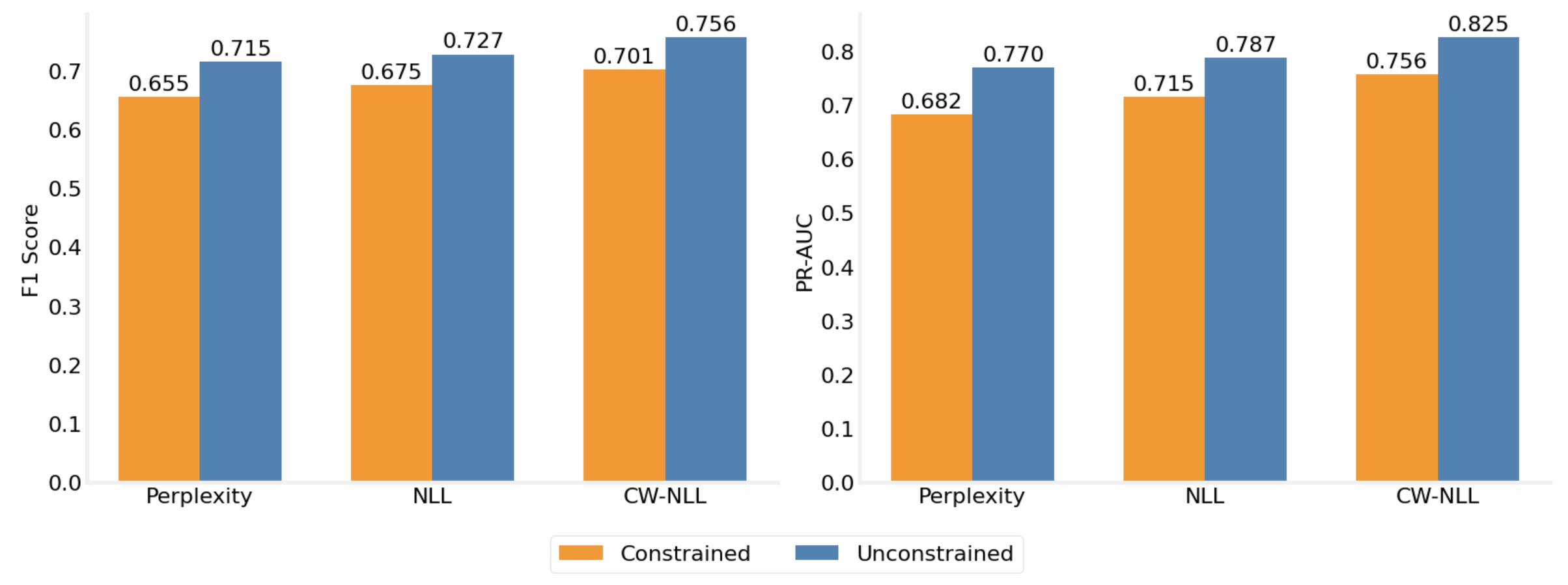}
  \caption{
  Comparison of scoring strategies on the Porto dataset. Each bar shows the F1 score (left) or PR-AUC (right) achieved by the same trained GETAD model under three scoring methods: perplexity, negative log likelihood (NLL), and confidence-weighted NLL (CW-NLL). CW-NLL consistently outperforms the other methods.
  }
  \label{fig:score_comparison}
  \Description{}
\end{figure*}

\vspace{-15pt}
\subsection{Porto Dataset Performance}

GETAD achieves the highest performance across both anomaly types and metrics. For constrained anomalies, it records the best F1 score of 0.701 and the best PR-AUC of 0.756. LM-TAD follows closely with an F1 score of 0.695 and a PR AUC of 0.732. GM-VSAE ranks third, achieving an F1 of 0.677 and a PR AUC of 0.664.

In the unconstrained setting, where anomalies are allowed to diverge more significantly from the original path, GETAD again achieves the highest PR-AUC of 0.825 and ties with GM-VSAE for the second-best F1 score of 0.756. GM VSAE exhibits a strong recovery in this setting, attaining the highest F1 score overall of 0.769 and the second-highest PR-AUC of 0.796, slightly outperforming LM-TAD.

These results reveal a clear trend: all models perform better in the unconstrained setting, with GM-VSAE in particular benefiting significantly from the increased anomaly magnitude. The performance of simpler models, such as SAE and VSAE, is notably lower across both anomaly types and metrics, especially in terms of PR AUC, indicating limited ability to capture fine-grained or rare trajectory deviations. The constrained anomaly setting presents greater difficulty due to the proximity of anomalous and normal paths. These anomalies are designed to represent slight deviations, often only a few road segments away from the expected route, which poses a major challenge for models that do not explicitly capture spatial structure. Despite this difficulty, GETAD maintains strong performance, outperforming all baselines. Its advantage is most evident in PR-AUC, where its margin over the next best model (LM-TAD) is 2.4 percentage points.

The strong performance of GETAD in both constrained and unconstrained settings stems from its integration of spatially aware segment embeddings, graph-informed positional encoding, and a confidence-weighted scoring mechanism. These components enable it to distinguish between behaviorally plausible variations and structurally implausible deviations, even when the observed anomaly is minor.
\section{Ablation Study}
\subsection{Anomaly Scoring Comparison}

We evaluate the impact of different anomaly scoring strategies on the performance of GETAD using the Porto dataset. Specifically, we compare the proposed Confidence Weighted Negative Log Likelihood (CW-NLL) with two widely used alternatives: standard Negative Log Likelihood (NLL) and Perplexity.

All three scores are derived from the same trained model and differ only in how they aggregate prediction probabilities across the trajectory. While NLL simply sums the token-level log probabilities, Perplexity takes the exponentiated average and has been commonly used in language modeling. CW-NLL builds on NLL by incorporating the model’s prediction confidence at each step, reducing the influence of uncertain predictions and emphasizing unlikely outcomes made with high certainty.

Figure~\ref{fig:score_comparison} shows the F1 and PR-AUC results under both constrained and unconstrained anomaly settings. Across all scoring strategies, CW-NLL consistently achieves the highest performance. In the constrained setting, CW-NLL improves PR-AUC from 0.682 (Perplexity) and 0.715 (NLL) to 0.756. In the unconstrained setting, it further raises PR-AUC to 0.825, outperforming NLL (0.787) and Perplexity (0.770). These results highlight the importance of accounting for model confidence when scoring anomalies. By emphasizing high-certainty, low-likelihood predictions, CW-NLL produces a more focused and reliable anomaly signal, especially in real-world settings where uncertainty is common and multiple valid continuations may exist.

\subsection{Interaction Between Positional Encoding and Structural Supervision}

\begin{figure}[t]
  \centering
  \includegraphics[width=0.45\textwidth]{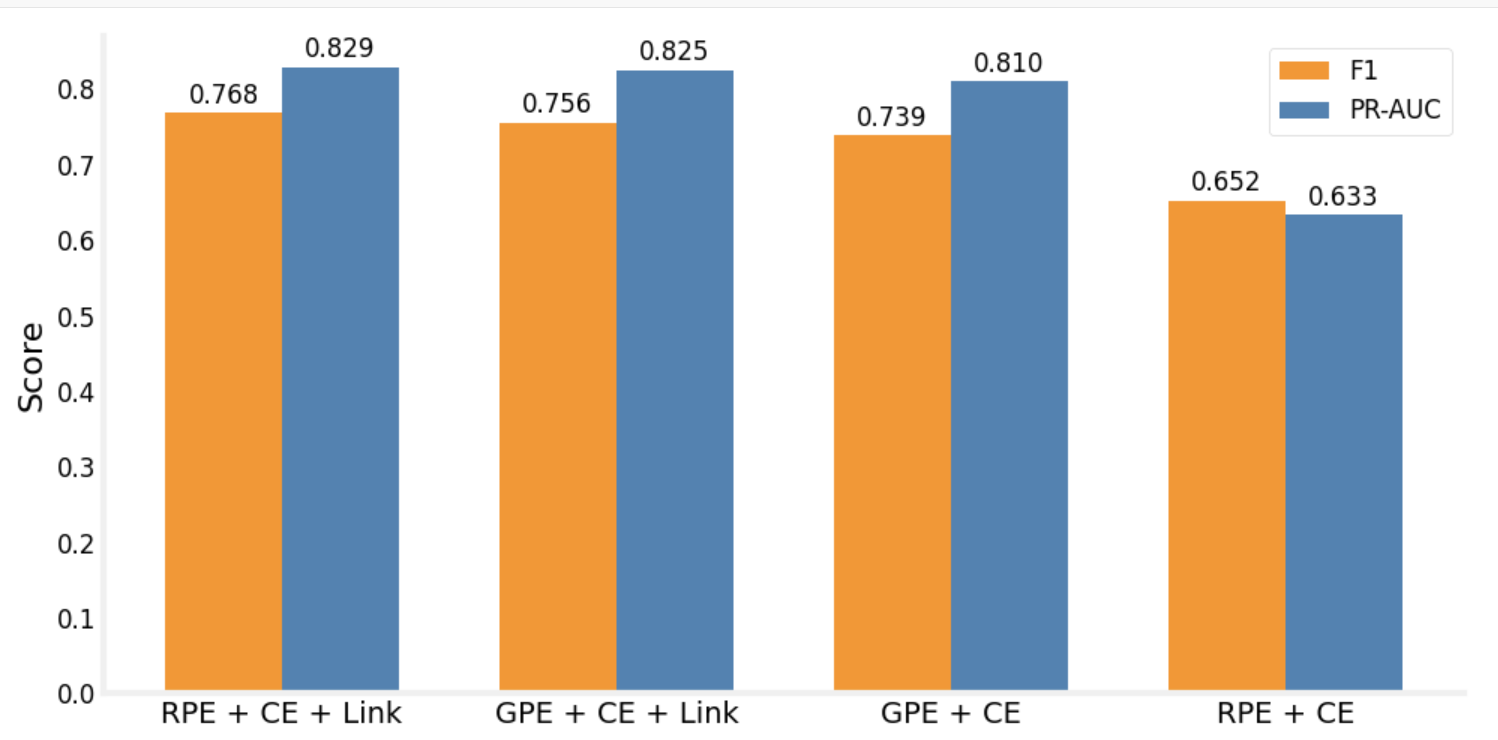}
  \caption{Performance comparison of positional encoding strategies with and without link prediction (Link) loss on the Porto dataset. Relative positional encoding (RPE) combined with link prediction yields the highest F1 and PR-AUC. Graph positional encoding (GPE) improves performance over relative encoding when no structural supervision is provided, but does not offer additional benefit when the link prediction objective is used.}

  \label{fig:positional_ablation}
  \Description{}
\end{figure}

We investigate how different positional encoding strategies interact with structural supervision in shaping model performance. Figure ~\ref{fig:positional_ablation} summarizes results from four configurations: Graph Positional Encoding (GPE) or Relative Positional Encoding (RPE), each trained with or without the link prediction loss. All models use cross-entropy loss for autoregressive trajectory prediction.

The results reveal that the model requires an explicit mechanism for learning proximity between road segments in order to detect subtle detour anomalies. When using RPE without structural supervision, performance is substantially lower (0.633 F1 and 0.652 PR-AUC), indicating the model's difficulty in capturing spatial relationships implicitly. In contrast, adding link prediction loss to RPE dramatically improves results to 0.768 F1 and 0.829 PR-AUC. This suggests that the link loss encourages the model to learn meaningful topological relationships between segments that are essential for accurate detection.

GPE also encodes spatial proximity by design. Even without link supervision, the GPE-based model achieves 0.739 F1 and 0.810 PR-AUC, significantly outperforming the unsupervised RPE baseline. However, when link prediction is added, the RPE variant slightly outperforms GPE (0.829 vs. 0.825 PR-AUC), indicating that GPE and link supervision provide partially overlapping benefits.

These findings highlight that proximity information is critical to model performance. Whether injected through graph-based encodings or learned via structural supervision, the ability to reason about spatial connectivity between segments enables the model to move beyond simple sequence learning and capture deviations in road-constrained movement. Although RPE with link loss achieves the highest scores, GPE remains a valuable design choice because it encodes spatial proximity directly from the graph structure, without relying on additional supervision objectives during training.

\section{Conclusion} \label{sec:conclusion}
This work introduces GETAD, a graph-enhanced framework for trajectory anomaly detection that integrates road network structure, segment semantics, and historical movement patterns into a unified modeling pipeline. By treating map-matched trajectories as sequences over a structured graph, GETAD effectively captures both spatial and sequential dependencies critical for distinguishing between typical and anomalous behavior in road-constrained environments.

Through comprehensive experiments on the synthetic Simulation dataset and the Porto taxi dataset, we demonstrated that GETAD consistently outperforms existing baselines, especially in detecting subtle detours and irregular transitions that are otherwise difficult to flag using frequency-based or purely Euclidean methods. The combination of Graph Attention Networks and graph-based positional encodings enables the model to reason about road connectivity, segment type, and transition likelihood when evaluating trajectory behavior.

We also introduced Confidence-Weighted Negative Log Likelihood (CW-NLL), a novel scoring function that prioritizes low-probability predictions made with high model confidence. This improves the model’s ability to distinguish between truly anomalous events and rare but plausible variations, reducing false positives without sacrificing sensitivity.


GETAD provides a \textit{graph-aware}, \textit{context-sensitive}, and \textit{efficient} solution for trajectory anomaly detection. By modeling road structure and movement behavior jointly, it offers a principled approach for detecting irregularities in structured mobility data.
Although our evaluation is limited to detour anomalies, the design of GETAD is generalizable. By modeling road connectivity and sequential dependencies, the framework can extend to other anomaly types such as irregular transitions (e.g., erratic driving, sudden reversals) or temporal deviations (e.g., prolonged stops, off-schedule behavior) when additional contextual features are incorporated. We leave this extension to future work.”

\section*{Acknowledgment}

This work was supported by the National Science Foundation (Award 2127901) and by the Intelligence Advanced Research Projects Activity (IARPA) via Department of Interior/ Interior Business Center (DOI/IBC) contract number 140D0419C0050.  The U.S. Government is authorized to reproduce and distribute reprints for Governmental purposes, notwithstanding any copyright annotation thereon. Disclaimer: The views and conclusions contained herein are those of the authors and should not be interpreted as necessarily representing the official policies or endorsements, either expressed or implied, of IARPA, DOI/IBC, or the U.S. Government. Additionally, this work was supported by resources provided by the Office of Research Computing, George Mason University and by the National Science Foundation (Award Numbers 1625039, 2018631). 

\newpage
\bibliographystyle{ACM-Reference-Format}
\bibliography{main}




\end{document}